\documentclass[10pt,conference,a4paper]{IEEEtran}
\usepackage{dirtytalk}
\usepackage{subfig}
\usepackage{tabularx}
\newcolumntype{Y}{>{\centering\arraybackslash}X}

\ifCLASSINFOpdf
  \usepackage[pdftex]{graphicx}
\else
  \usepackage[dvips]{graphicx}
  
\fi

\usepackage{hyperref}

\begin{document}
\title{Adversarial Training for Aspect-Based Sentiment Analysis with BERT}

\author{\IEEEauthorblockN{Akbar Karimi}
\IEEEauthorblockA{IMP Lab - D.I.A.\\
University of Parma\\
Parma, Italy\\
Email: akbar.karimi@unipr.it
}
\and
\IEEEauthorblockN{Leonardo Rossi}
\IEEEauthorblockA{IMP Lab - D.I.A.\\
University of Parma\\
Parma, Italy\\
Email: leonardo.rossi@unipr.it}
\and
\IEEEauthorblockN{Andrea Prati}
\IEEEauthorblockA{IMP Lab - D.I.A.\\
University of Parma\\
Parma, Italy\\
Email: andrea.prati@unipr.it}
}


%Telephone: (800) 555--1212, Fax: (888) 555--1212}


\maketitle

\begin{abstract}
Aspect-Based Sentiment Analysis (ABSA) studies the extraction of sentiments and their targets. Collecting labeled data for this task in order to help neural networks generalize better can be laborious and time-consuming. As an alternative, similar data to the real-world examples can be produced artificially through an adversarial process which is carried out in the embedding space. Although these examples are not real sentences, they have been shown to act as a regularization method which can make neural networks more robust.  In this work, we fine-tune the general purpose BERT and domain specific post-trained BERT (BERT-PT) using adversarial training. After improving the results of post-trained BERT with different hyperparameters, we propose a novel architecture called BERT Adversarial Training (BAT) to utilize adversarial training for the two major tasks of Aspect Extraction and Aspect Sentiment Classification in sentiment analysis. The proposed model outperforms the general BERT as well as the in-domain post-trained BERT in both tasks. To the best of our knowledge, this is the first study on the application of adversarial training in ABSA.

The code is publicly available on a GitHub repository at \url{https://github.com/IMPLabUniPr/Adversarial-Training-for-ABSA}
\end{abstract}


% \fi
\IEEEpeerreviewmaketitle

\makeatletter{\renewcommand*{\@makefnmark}{}
	\footnotetext{© 2020 IEEE.  Personal use of this material is permitted.  Permission from IEEE must be obtained for all other uses, in any current or future media, including reprinting/republishing this material for advertising or promotional purposes, creating new collective works, for resale or redistribution to servers or lists, or reuse of any copyrighted component of this work in other works.}\makeatother

\section{Introduction}
Understanding what people are talking about and how they feel about it is valuable especially for industries which need to know the customers' opinions on their products. Aspect-Based Sentiment Analysis (ABSA) is a branch of sentiment analysis which deals with extracting the opinion targets (aspects) as well as the sentiment expressed towards them. For instance, in the sentence \say{The spaghetti was out of this world.}, a positive sentiment is mentioned towards the target which is \say{spaghetti}. Performing these tasks requires a deep understanding of the language. Traditional machine learning methods such as Support Vector Machines (SVM) \cite{kiritchenko2014nrc}, Naive Bayes \cite{gamallo2014citius}, Decision Trees \cite{wakade2012text}, Maximum Entropy \cite{nigam1999using} have long been practiced to acquire such knowledge. However, in recent years due to the abundance of available data and computational power, deep learning methods such as Convolutional Neural Nets (CNNs) \cite{lecun1995convolutional, kim2014convolutional, zhang2015character}, Recurrent Neural Networks (RNNs) \cite{liu2015fine, wang2016attention, ma2018targeted}, and the Transformer \cite{vaswani2017attention} have outperformed the traditional machine learning techniques in various tasks of sentiment analysis. Bidirectional Encoder Representations from Transformers (BERT) \cite{devlin2019bert} is a deep and powerful language model which uses the encoder of the Transformer in a self-supervised manner to learn the language model. It has been shown to result in state-of-the-art performances on the GLUE benchmark \cite{wang2018glue} including text classification. It has been shown in \cite{xu2019bert} that adding domain-specific information to this model can enhance its performance in ABSA. Using their post-trained BERT (BERT-PT), we add adversarial examples to further improve BERT's performance on Aspect Extraction (AE) and Aspect Sentiment Classification (ASC) which are two major tasks in ABSA. A brief overview of these two sub-tasks is given in Section \ref{absa}.

Adversarial examples are a way of fooling a neural network to behave incorrectly \cite{szegedy2014intriguing}. They are created by applying small perturbations to the original inputs. In the case of images, the perturbations can be invisible to human eye, but can cause neural networks to output a completely different response from the true one. Since neural nets make mistakes on these examples, introducing them to the network during the training can improve their performance. This is called \say{adversarial training} which acts as a regularizer to help the network generalize better \cite{goodfellow2014explaining}. Due to the discrete nature of text, it is not feasible to produce perturbed examples from the original inputs. As a workaround, \cite{miyato2016adversarial} apply this technique to the word embedding space for text classification. Inspired by them and building on the work of \cite{xu2019bert}, we experiment with adversarial training for ABSA.

Our main contribution in this work is the proposal of a novel architecture to apply adversarial training in the fine-tuning process of BERT language model for aspect extraction and aspect sentiment classification tasks in sentiment analysis. Our experiments show that the proposed model outperforms the performances of general BERT as well as domain specific post-trained BERT (BERT-PT) in general. As a minor contribution we demonstrate that the number of training epochs and dropout values can have significant impacts on the model's performance.  

\section{Related Work}\label{sec:related}

Since the early works on ABSA \cite{hu2004mining, titov2008joint, thet2010aspect}, several methods have been put forward to address the problem. In this section, we review some of the works which have utilized deep learning techniques. 

In \cite{poria2016aspect} the authors design a seven-layer CNN architecture and make use of both part of speech tagging and word embeddings as features. In \cite{xu2018double}, convolutional neural networks and domain-specific data are utilized for AE and ASC. They show that adding the word embeddings produced from the domain-specific data to the general purpose embeddings semantically enriches them regarding the task at hand. In a recent work \cite{xu2019bert}, the authors also show that using in-domain data can enhance the performance of the state-of-the-art language model (BERT). Similarly, \cite{rietzler2020adapt} also fine-tune BERT on domain-specific data for ASC. They perform a two-stage process, first of which is self-supervised in-domain fine-tuning, followed by supervised task-specific fine-tuning. Working on the same task, \cite{zhao2020modeling} apply graph convolutional networks taking into consideration the assumption that in sentences with multiple aspects, the sentiment about one aspect can help determine the sentiment of another aspect. 

Since its introduction by \cite{bahdanau2014neural}, attention mechanism has become widely popular in many natural language processing tasks including sentiment analysis. In \cite{li2019exploiting}, the authors design a network to transfer aspect knowledge learned from a coarse-grained network which performs aspect category sentiment classification to a fine-grained one performing aspect-level sentiment classification. This is carried out using an attention mechanism (Coarse2Fine) which contains an autoencoder that emphasizes the aspect term by learning its representation from the category embedding. Similar to the Transformer, which does away with RNNs and CNNs and use only attention for translation, \cite{song2019attentional} design an attention model for ASC with the difference that they use lighter (weight-wise) multi-head attentions for context and target word modeling.  
Using bidirectional LSTMs \cite{hochreiter1997long}, \cite{li2018aspect} propose a model that takes into account the history of aspects with an attention block called Truncated History Attention (THA). To capture the opinion summary, they also introduce Selective Transformation Network (STN) which highlights more important information with respect to a given aspect. In \cite{he2017unsupervised}, aspect extraction task is approached in an unsupervised way. Functioning the same way as an autoencoder, their model has been designed to reconstruct sentence embeddings in which aspect-related words are given higher weights through attention mechanism. 

While adversarial training has been utilized for sentence classification \cite{miyato2016adversarial, zhu2019freelb}, its effects have not been studied in ABSA. Therefore, in this work, we study the impact of applying adversarial training to the powerful BERT language model.

\section{Aspect-Based Sentiment Analysis Tasks} \label{absa}

In this section, we give a brief description of two major tasks in ABSA which are called Aspect Extraction (AE) and Aspect Sentiment Classification (ASC). These tasks were sub-tasks of task 4 in SemEval 2014 contest \cite{pontiki-etal-2014-semeval}, and since then they have been the focus of attention in many studies.

\textbf{Aspect Extraction.}
Given a collection of review sentences, the goal is to extract all the terms, such as \say{waiter}, \say{food}, and \say{price} in the case of restaurants, which point to aspects of a larger entity \cite{pontiki-etal-2014-semeval}. In order to perform this task, it is usually modeled as a sequence labeling task, where each word of the input is labeled as one of the three letters in \{B, I, O\}. Label \lq B' stands for \say{Beginning} of the aspect terms, \lq I' for \say{Inside} (aspect terms' continuation), and \lq O' for \say{Outside} or non-aspect terms. The reason for \say{Inside} label is that sometimes aspects can contain two or more words and the system has to return all of them as the aspect. In order for a sequence ($s$) of $n$ words to be fed into the BERT architecture, they are represented as 
\begin{center}
	$[CLS], w_1, w_2, ..., w_n, [SEP]$
\end{center}
where the $[CLS]$ token is an indicator of the beginning of the sequence as well as its sentiment when performing sentiment classification. The $[SEP]$ token is a token to separate a sequence from the subsequent one. Finally, $w_{i}$ are the words of the sequence. After they go through the BERT model, for each item of the sequence, a vector representation of the size 768, size of BERT's hidden layers, is computed. Then, we apply a fully connected layer to classify each word vector as one of the three labels.

\textbf{Aspect Sentiment Classification.}
Given the aspects with the review sentence, the aim in ASC is to classify the sentiment towards each aspect as {Positive, Negative, Neutral}. In this task, the input format for the BERT model is the same as in AE. The $[CLS]$ token in the input representation (Figure \ref{bertemb})  of the BERT is where the sentiment is encoded. After the input goes through the network, in the last layer the sentiment is extracted from this token by applying a fully connected layer to its encoding. 

Input sequences can have multiple aspects, meaning that a sequence can contain multiple targets with a specific sentiment polarity while BERT architecture has one element which is responsible for sentiment representation of each input. This problem is addressed in the preprocessing step when preparing the input data for the model. Sequences with multiple aspects are repeated as many times as the number of aspects they contain, each time with one of their specific aspect sentiments. 

\section{Model}\label{model}

\begin{figure}
	\includegraphics[scale=0.2]{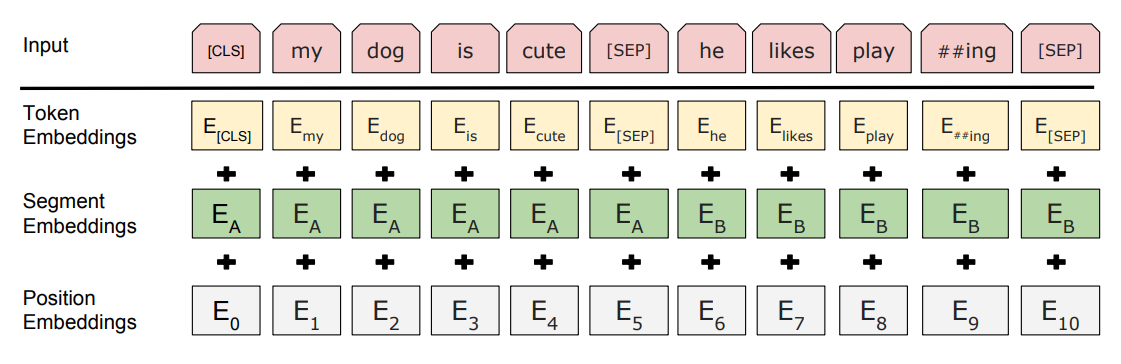}
	\caption{BERT word embedding layer \cite{devlin2019bert}}
	\label{bertemb}
\end{figure}

\begin{figure}
	\centering
	\includegraphics[scale=0.22]{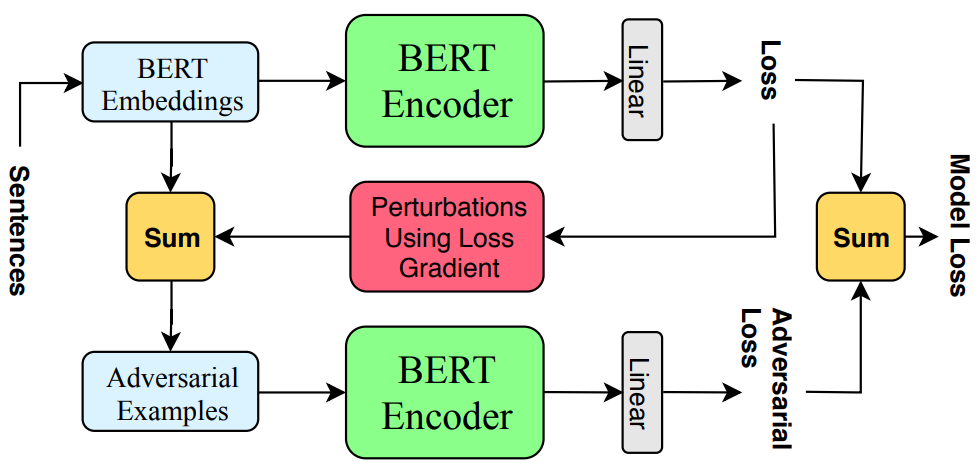}
	\caption{The proposed architecture: BERT Adversarial Training (BAT)}
	\label{bat_model}
\end{figure}

Our model is depicted in Figure \ref{bat_model}. As can be seen, we create adversarial examples from BERT embeddings using the gradient of the loss. Then, we feed the perturbed examples to the BERT encoder to calculate the adversarial loss. In the end, the backpropagation algorithm is applied to the sum of both losses. 

\textbf{BERT Word Embedding Layer.} The calculation of input embeddings in BERT is carried out using three different embeddings. As shown in Figure \ref{bertemb}, it is computed by summing over token, segment, and position embeddings. Token embedding is the vector representation of each token in the vocabulary which is achieved using WordPiece embeddings \cite{wu2016google}. Position embeddings are used to preserve the information about the position of the words in the sentence. Segment embeddings are used in order to distinguish between sentences if there is more than one (e.g. for question answering task there are two). Words belonging to one sentence are labeled the same. 

\textbf{BERT Encoder.} 
BERT encoder is constructed by making use of Transformer blocks from the Transformer model. For BERT-BASE, these blocks are used in 12 layers, each of which consists of 12 multi-head attention blocks. In order to make the model aware of both previous and future contexts, BERT uses the Masked Language Model (MLM) where $15\%$ of the input sentence is masked for prediction. 

\textbf{Fully Connected Layer and Loss Function.} The job of the fully connected layer in the architecture is to classify the output embeddings of BERT encoder into sentiment classes. Therefore, its size is $768\times3$ where the first element is the hidden layers' size of BERT encoder and the second element is the number of classes. For the loss function, we use cross entropy loss implemented in Pytorch. 

\textbf{Adversarial Examples.} Adversarial examples are created to attack a neural network to make erroneous predictions. There are two main types of adversarial attacks which are called white-box and black-box. White-box attacks \cite{ebrahimi2018hotflip} have access to the model parameters, while black-box attacks \cite{ilyas2018black} work only on the input and output. In this work, we utilize a white-box method working on the embedding level. In order to create adversarial examples, we utilize the formula used by \cite{miyato2016adversarial}, where the perturbations are created using gradient of the loss function. Assuming $p(y|x;\theta)$ is the probability of label $y$ given the input $x$ and the model parameters $\theta$, in order to find the adversarial examples the following minimization problem should be solved:
\begin{equation}
r_{adv} = \smash{\displaystyle \arg \min_{r, ||r|| \leq \epsilon}} \log\, p(y|x+r;\hat{\theta})
\label{radv}
\end{equation}
where $r$ denotes the perturbations on the input and $\hat{\theta}$ is a constant copy of $\theta$ in order not to allow the gradients to propagate in the process of constructing the artificial examples. Solving the above minimization problem means that we are searching for the worst perturbations while trying to minimize the loss of the model. An approximate solution for Equation \ref{radv} is found by linearizing $\log p(y|x;\theta)$ around $x$ \cite{goodfellow2014explaining}. Therefore, the following perturbations are added to the input embeddings to create new adversarial sentences in the embedding space. 

\begin{equation}
r_{adv} = -\epsilon \frac{g}{||g||_2}
\end{equation} 
where 
\begin{equation}
g = \nabla_{x}\, \log\,  p(y|x;\hat{\theta})
\end{equation} and $\epsilon$ is the size of the perturbations. In order to find values which outperform the original results, we carried out experiments with five values for epsilon whose results are presented in Figure \ref{ae_adv} and discussed in Section \ref{experiments}. After the adversarial examples go through the network, their loss is calculated as follows: 
\begin{center}
	$- \log p(y|x + r_{adv};\theta)$
\end{center}

Then, this loss is added to the loss of the real examples in order to compute the model's overall loss.
\section{Experimental Setup}\label{setup}

\textbf{Datasets.} In order for the results to be consistent with previous works, we experimented with the benchmark datasets from SemEval 2014 task 4 \cite{pontiki-etal-2014-semeval} and SemEval 2016 task 5 \cite{pontiki2016semeval} competitions. The laptop dataset is taken from SemEval 2014 and is used for both AE and ASC tasks. However, the restaurant dataset for AE is a SemEval 2014 dataset while for ASC is a SemEval 2016 dataset. The reason for the difference is to be consistent with BERT-PT. A summary of these datasets can be seen in Tables \ref{tab:ae} and \ref{tab:asc}.

\begin{table}
	\caption{Laptop and restaurant datasets for AE. S: Sentences; A: Aspects; Rest16: Restaurant dataset from SemEval 2016}
	\begin{center}
		\begin{tabularx}{\columnwidth}{@{}l|Y|Y|Y|Y@{}} \hline
			& \multicolumn{2}{c|}{Train} & \multicolumn{2}{c}{Test}\\
			\hline
			\textbf{Dataset} & S & A & S & A\\
			\hline
			Laptop & 3045 & 2358 & 800 & 654 \\
			\hline
			Rest16 & 2000 & 1743 & 676 & 622\\
			\hline
		\end{tabularx}
	\end{center}
	\label{tab:ae}
\end{table}

\begin{table}
	\caption{Laptop and restaurant datasets for ASC. Pos, Neg, Neu: Number of positive, negative, and neutral sentiments, respectively; Rest14: Restaurant dataset from SemEval 2014}
	\begin{center}
		\begin{tabularx}{\columnwidth}{@{}l|Y|Y|Y|Y|Y|Y@{}} \hline
			& 
			\multicolumn{3}{c|}{Train}
			&
			\multicolumn{3}{c}{Test}\\
			\hline
			\textbf{Dataset} & Pos & Neg & Neu & Pos & Neg & Neu\\
			\hline
			Laptop & 987 & 866 & 460 & 341 & 128 & 169 \\
			\hline 
			Rest14 & 2164 & 805 & 633 & 728 & 196 & 196 \\
			\hline
		\end{tabularx}
	\end{center}
	\label{tab:asc}
\end{table} 

\textbf{Implementation details.} We performed all our experiments on a GPU (GeForce RTX 2070) with 8 GB of memory. Except for the code specific to our model, we adapted the codebase utilized by BERT-PT. For the Subsection A of Section \ref{experiments}, to carry out the experiments of BERT-PT model, batches of 32 were specified. However, for our proposed model, we reduced the batch size to 16 in order for the GPU to be able to store the network. For Subsection B, we used batches of 16 for all them. For optimization, the Adam optimizer with a learning rate of $3e-5$ was used. From SemEval's training data, 150 examples were chosen for the validation and the remaining was used for training the model. 

\begin{figure}
	\subfloat[Laptop]{\includegraphics[scale=0.45]{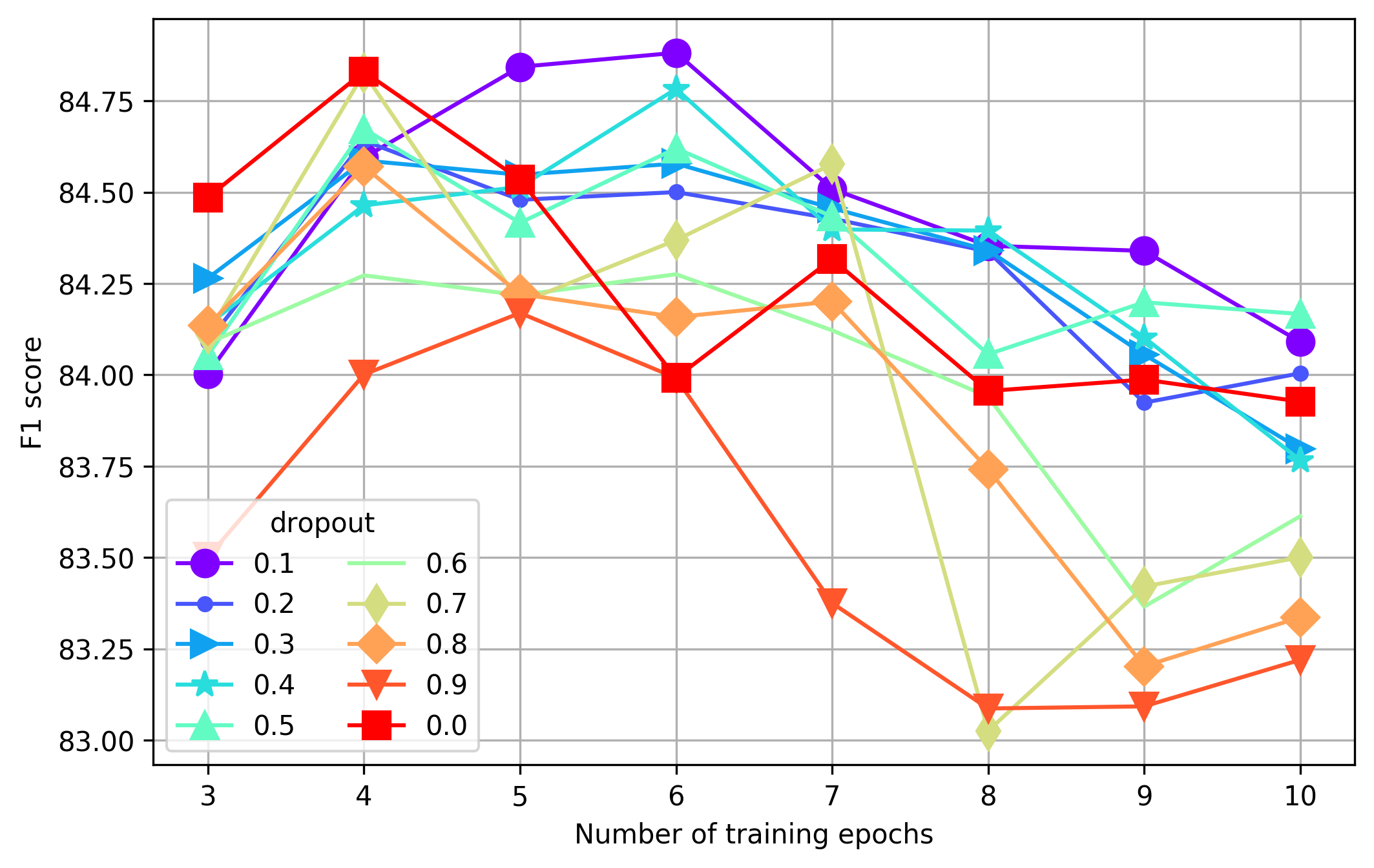}}
	
	\subfloat[Rest16]{\includegraphics[scale=0.45]{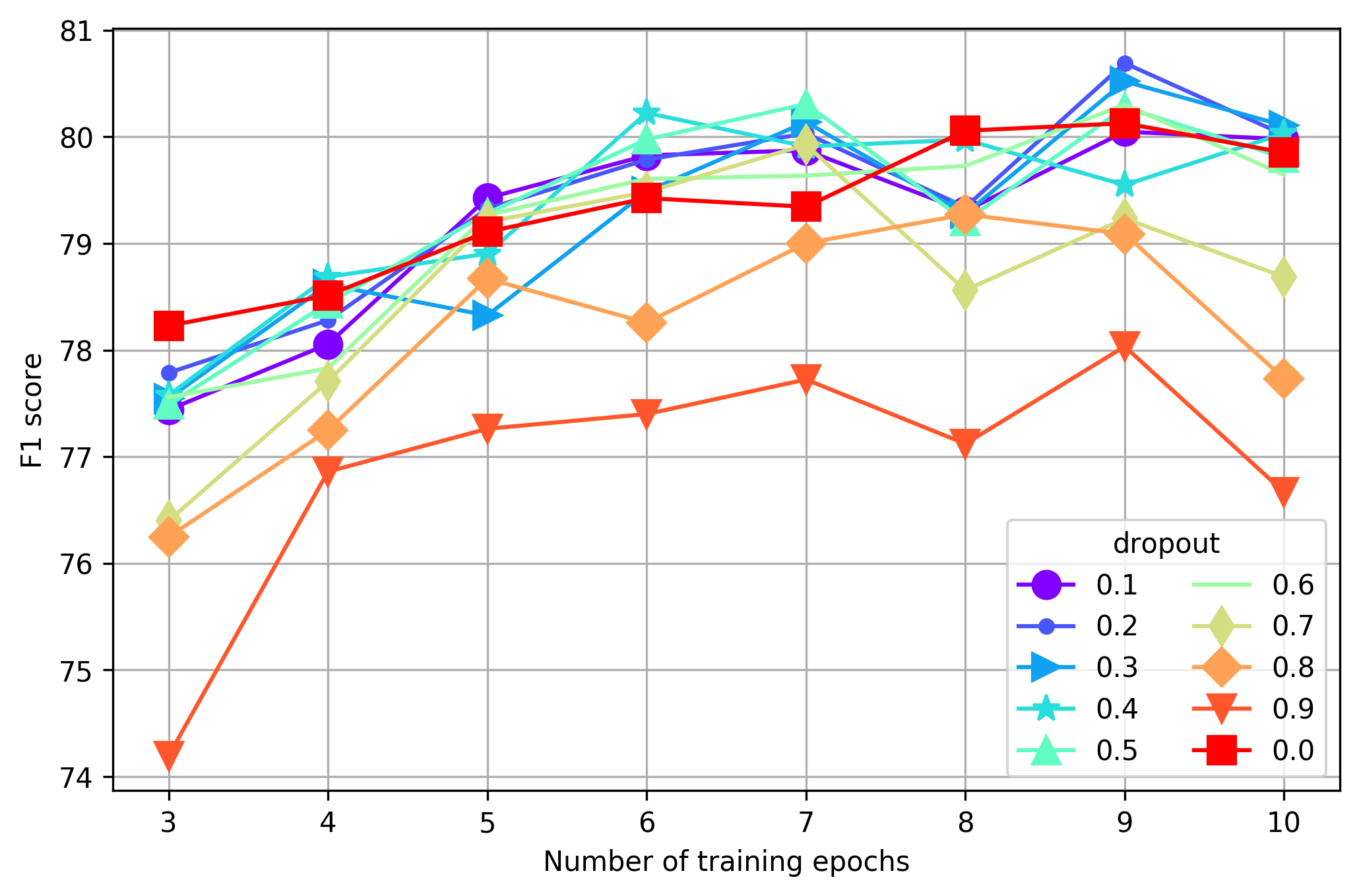}} 
	\caption{Results on the impact of training epochs and dropout value in post-trained BERT for AE task}
	\label{ae_pt}
\end{figure}
\begin{figure}
	\subfloat[Laptop]{\includegraphics[scale=0.45]{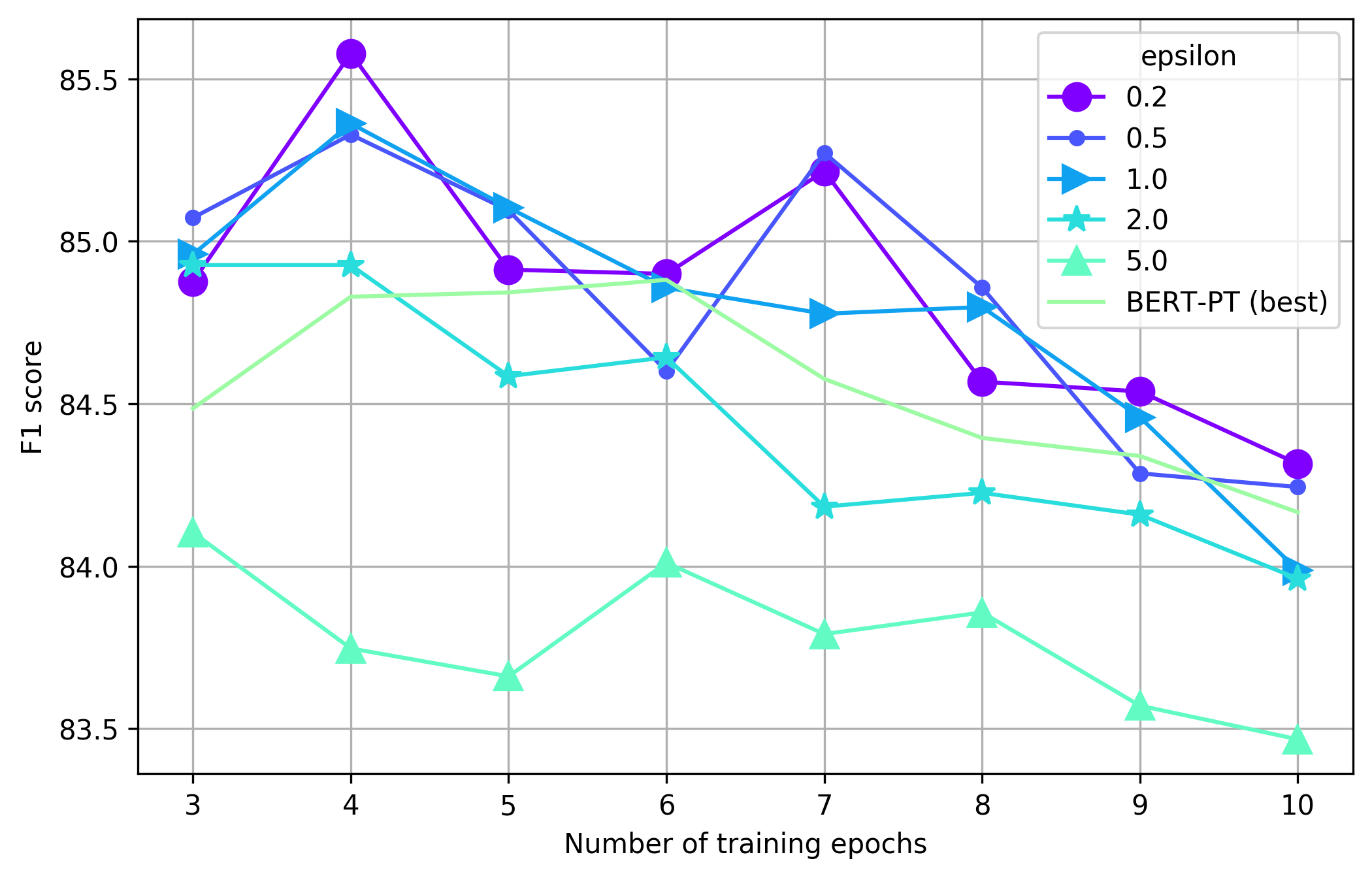}}
	
	\subfloat[Rest16]{\includegraphics[scale=0.45]{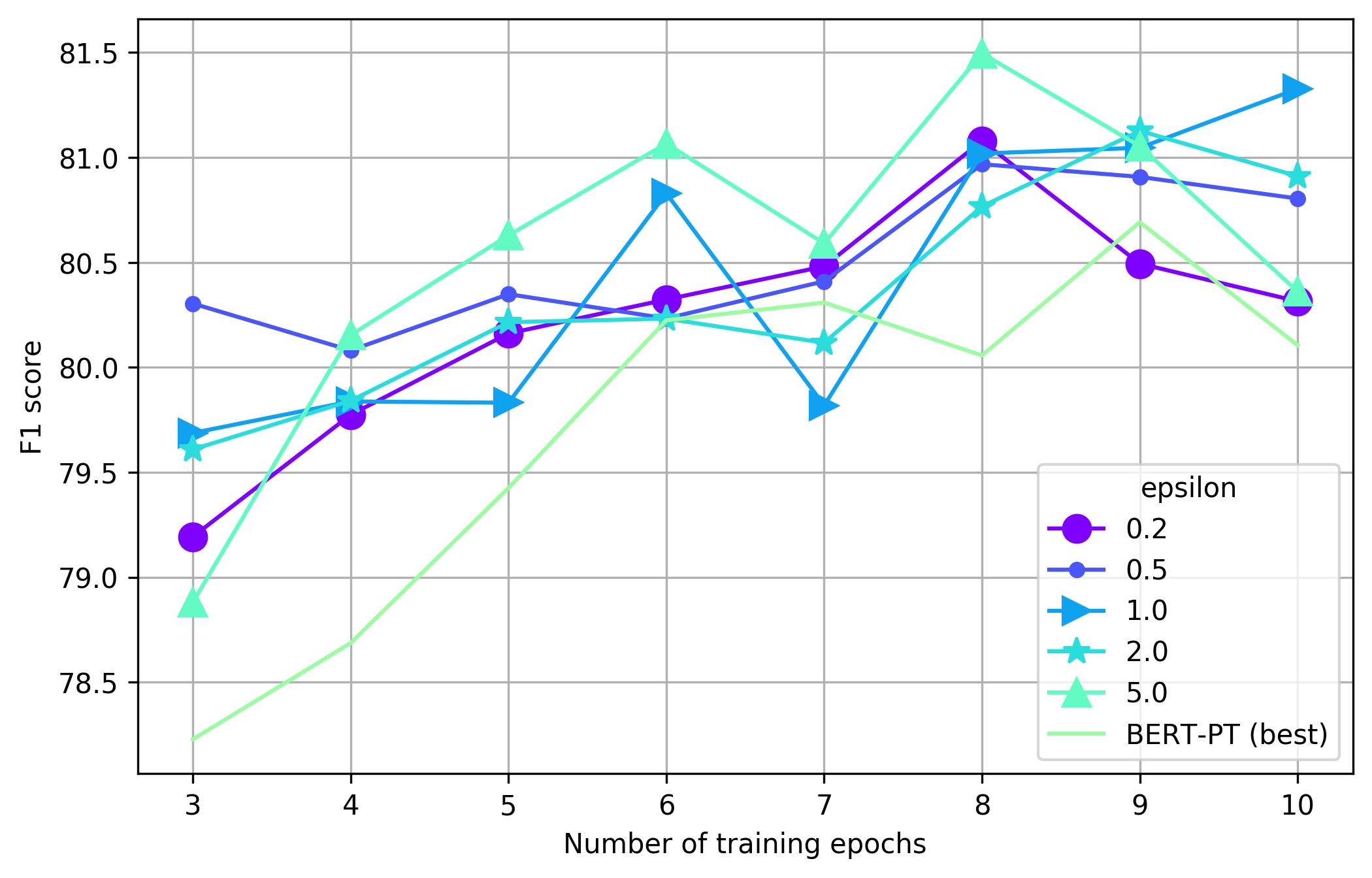}}
	\caption{Comparing best results of BERT-PT and BAT with different sizes of perturbations ($\epsilon$) for AE task}
	\label{ae_adv}
\end{figure}

Implementing the creation of adversarial examples for ASC task was slightly different from doing it for AE task. During our experiments, we realized that modifying all the elements of input vectors does not improve the results. Therefore, we decided not to modify the vector for the $[CLS]$ token. Since the $[CLS]$ token is responsible for the class label in the output, it seems reasonable not to change it in the first place and only perform the modification on the word vectors of the input sentence. In other words, regarding the fact that the $[CLS]$ token is the class label, to create an adversarial example, we should only change the words of the sentence, not the label.  

\textbf{Evaluation.} To evaluate the performance of the model, we utilized the official script of the SemEval contest for AE. These results are reported as F1 scores. For ASC, to be consistent with BERT-PT, we utilized their script whose results are reported in Accuracy and Macro-F1 (MF1) measures. Macro-F1 is the average of F1 score for each class and it is used to deal with the issue of unbalanced classes.

\begin{figure}[t]
	\subfloat[Laptop]{\includegraphics[scale=0.45]{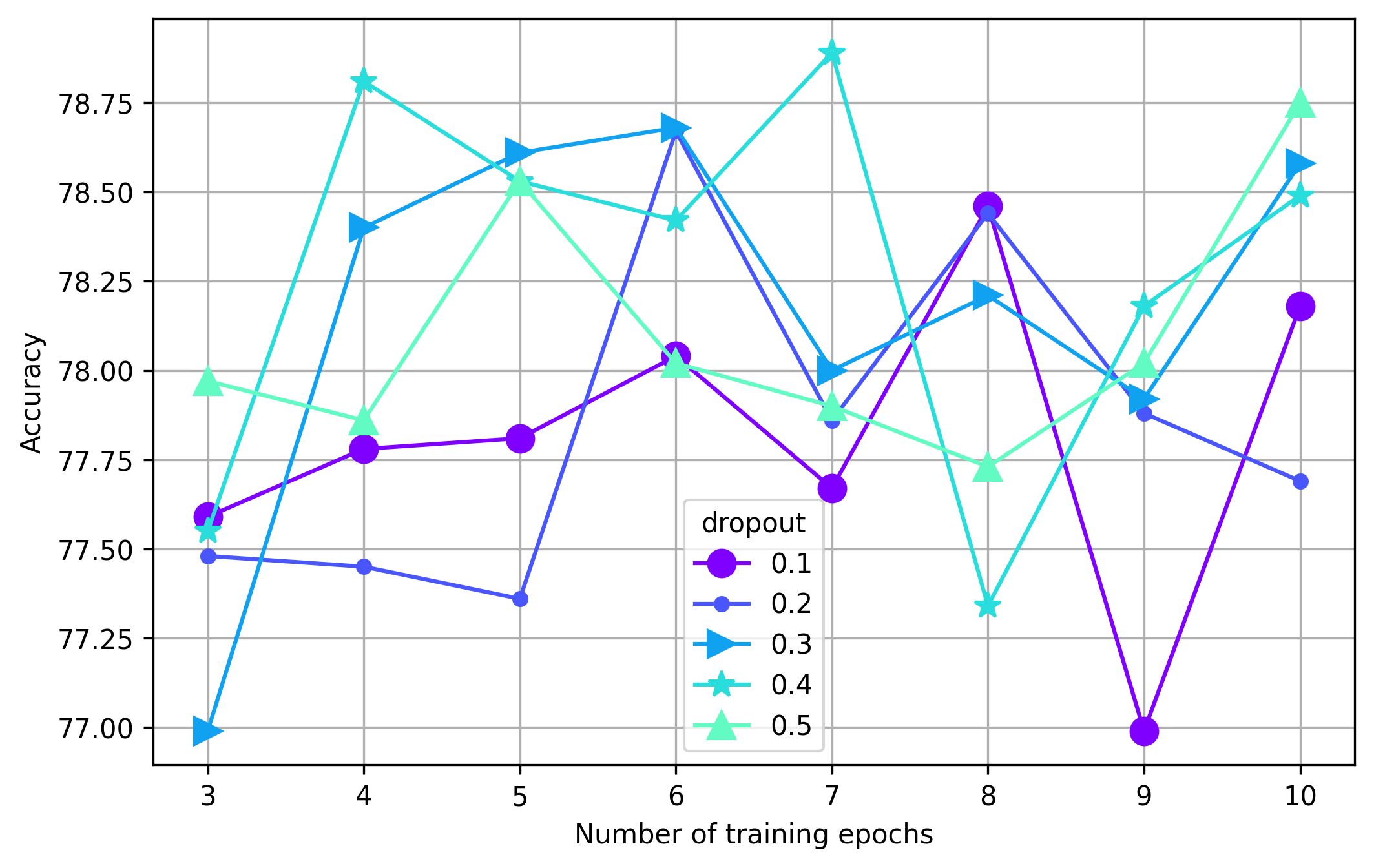}}
	
	\subfloat[Rest14]{\includegraphics[scale=0.45]{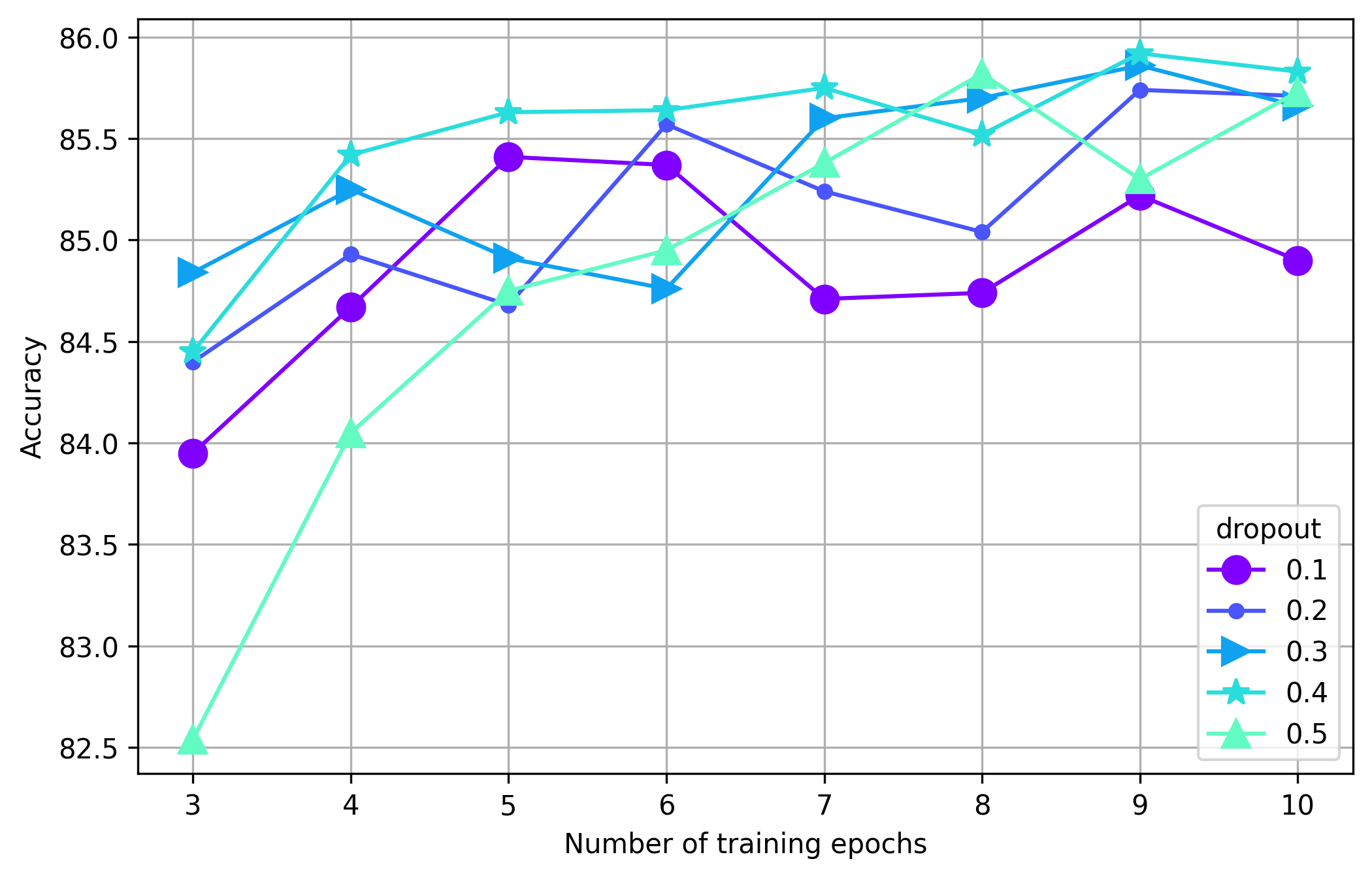}}
	\caption{Results on the impact of training epochs and dropout value in post-trained BERT for ASC task}
	\label{asc_pt}
\end{figure}

\begin{figure}[t]
	\subfloat[Laptop]{\includegraphics[scale=0.45]{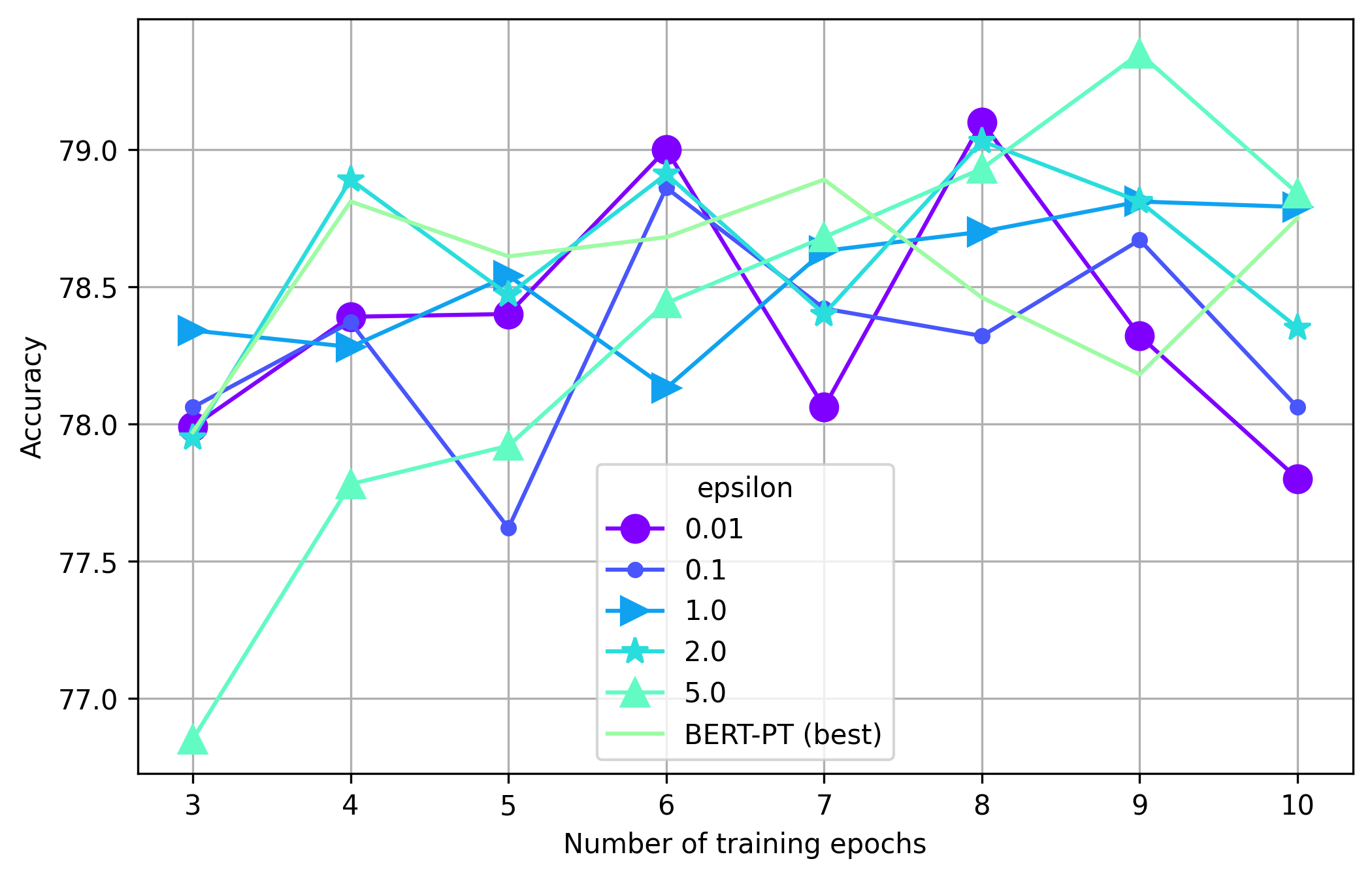}}
	
	\subfloat[Rest14]{\includegraphics[scale=0.45]{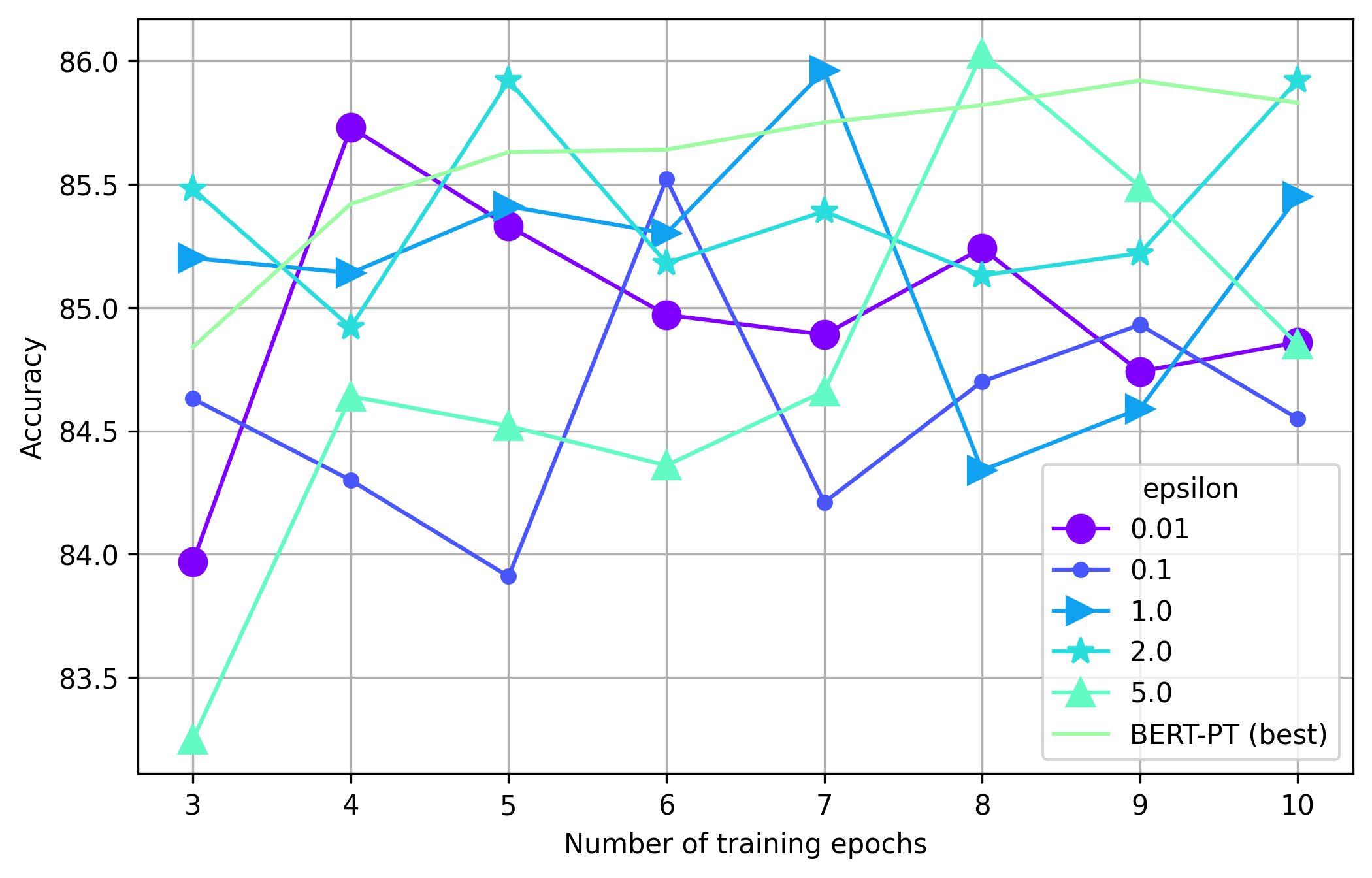}}
	\caption{Comparing best results of BERT-PT and BAT with different sizes of perturbations ($\epsilon$) for ASC task}
	\label{asc_adv}
\end{figure}

\section{Experiments}  \label{experiments}
\subsection{Hyperparameters and adversarial training}
In order to carry out the experiments, first we initialize our model with post-trained BERT which has been trained using uncased version of BERT-BASE on laptop and restaurant data. We attempt to discover what number of training epochs and  which dropout probability yield the best performance for BERT-PT. Since one and two training epochs result in very low scores, results of 3 to 10 training epochs have been depicted. For AE, we experiment with 10 different dropout values in the fully connected (linear) layer. The results can be seen in Figure \ref{ae_pt} for laptop and restaurant datasets. To be consistent with the previous work and because of the results having high variance, each point in the figure (F1 score) is the average of 9 runs. In the end, for each number of training epochs, a dropout value, which outperforms the other values, is found. 
In our experiments, we noticed that the validation loss increases after 2 epochs as has been mentioned in the original paper. However, the test results do not follow the same pattern. Looking at the figures, it can be seen that as the number of training epochs increases, better results are produced in the restaurant domain while in the laptop domain the scores go down. This can be attributed to the selection of validation sets as for both domains the last 150 examples of the SemEval training set were selected. Therefore, it can be said that the examples in the validation and test sets for laptop have more similar patterns than those of restaurant dataset. To be consistent with BERT-PT, we performed the same selection.  

In order to compare the effect of adversarial examples on the performance of the model, we choose the best dropout for each number of epochs and experiment with five different values for epsilon (perturbation size). The results for laptop and restaurant can be seen in Figure \ref{ae_adv}. As is noticeable, in terms of scores, they follow the same pattern as the original ones. Although most of the epsilon values improve the results, it can be seen in Figure \ref{ae_adv} that not all of them will enhance the model's performance. In the case of $\epsilon=5.0$ for AE, while it boosts the performance in the restaurant domain for most of the training epochs, it negatively affects the performance in the laptop domain. The reason for this could be the creation of adversarial examples which are not similar to the original ones but are labeled the same. In other words, the new examples greatly differ from the original ones but are fed to the net as being similar, leading to the network's poorer performance.

\begin{table}
	\tabcolsep=0.11cm
	\caption{Comparison of best results for Aspect Extraction}
	\begin{center}
		\begin{tabularx}{\columnwidth}{@{}l|Y|Y@{}} \hline
			Domain & Laptop & Rest16\\
			\hline
			
			\textbf{Methods} & F1 & F1\\
			\hline
			THA/STN \cite{li2018aspect} & 79.52 & 73.61\\
			DE-CNN \cite{xu2018double} & 81.59 & 74.37\\
			BERT \cite{xu2019bert} & 79.28 & 74.10\\
			BERT-PT \cite{xu2019bert} & 84.26 & 77.97\\
			\hline
			\textbf{BERT-PT} (best) & \textbf{84.88} & \textbf{80.69} \\
			\textbf{BAT} (Ours) & \textbf{85.57} & \textbf{81.50} \\
			\hline
		\end{tabularx}
	\end{center}
	\label{tab:results}
	
	\tabcolsep=0.11cm
	\caption{Comparison of best results for Aspect sentiment classification. Acc: Accuracy; MF1: Macro-F1}
	\begin{center}
		\begin{tabularx}{\columnwidth}{@{}l|Y|Y|Y|Y@{}} \hline
			Domain & \multicolumn{2}{c|}{Laptop} & \multicolumn{2}{c}{Rest14} \\
			\hline
			
			\textbf{Methods} & Acc & MF1 & Acc & MF1 \\
			\hline
			\small MGAN \cite{li2018aspect} & 76.21 & 71.42 & 81.49 & 71.48\\
			\small BERT \cite{xu2019bert} & 75.29 & 71.91  & 81.54 & 71.94\\
			\small BERT-PT \cite{xu2019bert} & 78.07 & 75.08 & 84.95 & 76.96\\
			\hline
			\textbf{BERT-PT} (best) & \textbf{78.89} & \textbf{75.89} & \textbf{85.92} & \textbf{79.12} \\
			\textbf{BAT} (Ours) & \textbf{79.35} & \textbf{76.5} & \textbf{86.03} & \textbf{79.24} \\
			\hline
		\end{tabularx}
	\end{center}
	\label{tab:results_asc}
\end{table}

\begin{table*}
	\tabcolsep=0.11cm
	\caption{Comparison of BAT results using BERT-BASE initialization}
	\begin{center}
		\begin{tabularx}{0.8\textwidth}{@{}l|Y|Y|Y|Y|Y|Y@{}} \hline
			& \multicolumn{2}{c|}{Aspect Extraction (AE)} & \multicolumn{4}{c}{Aspect Sentiment Classification (ASC)} \\
			\hline
			Domain & Laptop & Rest16 & \multicolumn{2}{c|}{Laptop} & \multicolumn{2}{c}{Rest14} \\
			\hline
			\textbf{Methods} & F1 & F1 & Acc & MF1 & Acc & MF1 \\
			\hline
			BERT-BASE & 81.19 & 75.54 & 75.46 & 71.88 & 80.73 & 70.82\\
			\hline
			BAT ($\epsilon=0.01$) & \textbf{81.93 (+0.74)} & 76.16 (+0.62) & 75.34 (-0.12) & 71.86 (-0.02) & 81.5 (+0.77) & 72.25 (+1.43)\\
			BAT ($\epsilon=0.1$) & 81.33 (+0.14) & 75.87 (+0.33) & \textbf{75.88 (+0.54)} & \textbf{72.21 (+0.33)} & 80.77 (+0.04) & 70.94 (+0.12)\\
			BAT ($\epsilon=1.0$) & 81.55 (+0.36) & 76.37 (+0.83) & 74.89 (-0.45) & 71.3 (-0.58) & 81.2 (+0.47) & 71.84 (+1.02)\\
			BAT ($\epsilon=2.0$) & 81.58 (+0.39) & \textbf{76.38 (+0.84)} & 75.71 (+0.25) & 71.99 (+0.11) & \textbf{82.27 (+1.54)} & \textbf{73.7 (+2.88)}\\
			BAT ($\epsilon=5.0$) & 80.34 (-0.85) & 76.26 (+0.72) & 74.16 (-1.3) & 70.89 (-0.99) & 80.34 (-0.39) & 71.04 (+0.22)\\
			\hline

		\end{tabularx}
	\end{center}
	\label{results_bert_base}
\end{table*}

\begin{table*}
	\tabcolsep=0.11cm
	\caption{Comparison of BAT results using BERT-PT initialization}
	\begin{center}
		\begin{tabularx}{0.8\textwidth}{@{}l|Y|Y|Y|Y|Y|Y@{}} \hline
			& \multicolumn{2}{c|}{Aspect Extraction (AE)} & \multicolumn{4}{c}{Aspect Sentiment Classification (ASC)} \\
			\hline
			Domain & Laptop & Rest16 & \multicolumn{2}{c|}{Laptop} & \multicolumn{2}{c}{Rest14} \\
			\hline
			\textbf{Methods} & F1 & F1 & Acc & MF1 & Acc & MF1 \\
			\hline
			BERT-PT & 84.8 & 79.22 & 77.53 & 74.72 & 84.54 & 76.57\\
			\hline
			BAT ($\epsilon=0.01$) & \textbf{85.25 (+0.45)} & 80.15 (+0.93) & 77.99 (+0.46) & 74.96 (+0.24) & 84.8 (+0.26) & 77.12 (+0.55)\\
			BAT ($\epsilon=0.1$) & 85.17 (+0.37) & 79.42 (+0.2) & 77.99 (+0.46) & 74.98 (+0.26) & 84.65 (+0.11) & 76.94 (+0.37)\\
			BAT ($\epsilon=1.0$) & 85.06 (+0.26) & 79.8 (+0.58) & 78.18 (+0.65) & 75.31 (+0.59) & 85.29 (+0.75) & 77.86 (+1.29)\\
			BAT ($\epsilon=2.0$) & 84.69 (-0.11) & 79.89 (+0.67) & \textbf{78.42 (+0.89)} & \textbf{75.32 (+0.6)} & 85.21 (+0.67) & 77.75 (+1.18)\\
			BAT ($\epsilon=5.0$) & 83.76 (-1.04) & \textbf{80.2 (+0.98)} & 76.8 (-0.73) & 73.72 (-1.0) & \textbf{85.6 (+1.06)} & \textbf{78.4 (+1.83)}\\
			\hline
			
		\end{tabularx}
	\end{center}
	\label{results_bert_pt}
\end{table*}

Observing, from AE task, that higher dropouts perform poorly, we experiment with the 5 lower values for ASC task in BERT-PT experiments. In addition, for BAT experiments, two different values \{$0.01, 0.1$\} for epsilon are tested to make them more diverse. The results are depicted in Figures \ref{asc_pt} and \ref{asc_adv} for BERT-PT and BAT, respectively. While in AE, towards higher number of training epochs, there is an upward trend for restaurant and a downward trend for laptop, in ASC a clear pattern is not observed. Regarding the dropout, lower values ($0.1$ for laptop, $0.2$ for restaurant) yield the best results for BERT-PT in AE task, but in ASC a dropout probability of 0.4 results in top performance in both domains. 
The top performing epsilon value for both domains in ASC, as can be seen in Figure \ref{asc_adv}, is 5.0 which is the same as the best value for restaurant domain in AE task. This is different from the top performing $\epsilon = 0.2$ for laptop in AE task which was mentioned above. 

From the experiments on BERT-PT hyperparameters, we extract the best results of BERT-PT and compare them with those of BAT. These are summarized in Tables \ref{tab:results} and \ref{tab:results_asc} for aspect extraction and aspect sentiment classification, respectively. The base results reported are from the corresponding papers. As can be seen in Table \ref{tab:results}, the best parameters for BERT-PT have greatly improved its original performance on restaurant dataset (\textbf{+2.72}) compared to laptop (\textbf{+0.62}). Similar improvements can be seen in ASC results with an increase of \textbf{+2.16} in MF1 score for restaurant compared to \textbf{+0.81} for laptop which is due to the increase in the number of training epochs for restaurant domain since it exhibits better results with more training while the model reaches its peak performance for laptop domain in earlier training epochs. In addition, applying adversarial training improves the network's performance in both tasks, though at different rates. While for laptop there are similar improvements in both tasks (\textbf{+0.69} in AE, \textbf{+0.61} in ASC), for restaurant we observe different enhancements (\textbf{+0.81} in AE, \textbf{+0.12} in ASC). This could be attributed to the fact that these are two different datasets whereas the laptop dataset is the same for both tasks. Furthermore, the perturbation size plays an important role in performance of the system. By choosing the appropriate ones, as was shown, better results are achieved. 

\subsection{Perturbation size in adversarial training} In order to discover the effect of adversarial training on the performance of the model, first we fix the number of training epochs at $4$ and dropout value at $0.1$ as our base configuration. Then we apply the BAT model with five perturbation sizes of $\{0.01, 0.1, 1.0, 2.0, 5.0\}$ to both BERT-BASE and BERT-PT with the same configurations. The results can be found in Tables \ref{results_bert_base} and \ref{results_bert_pt}, respectively. Results are reported from our implementation. As can be seen from the results, in the majority of cases for all the epsilon values, the performance over the base models improves. However, in a few cases and for the same reason which was mentioned above, it falls bellow the baseline. Although the BAT model in the search for the worst case adversarial examples opts for larger values leading to top performances in most cases, it runs the risk of breaking out of the safe zone and produces examples which are not similar to (or in the vicinity of) original ones. This causes the model to sometimes perform poorly with large epsilon values. On the other hand, the smallest values in general enhance the performance of both baseline models more consistently, though with less significant amounts.

\section{Conclusion}
In this paper, we introduced the application of adversarial training in Aspect-Based Sentiment Analysis. The experiments with our proposed architecture show that the performance of the general purpose BERT and in-domain post-trained BERT on aspect extraction and aspect sentiment classification tasks are improved by utilizing adversarial examples during the network training. As future work, other white-box adversarial attacks as well as black-box ones will be utilized for a comparison of adversarial training methods for various sentiment analysis tasks. Furthermore, the impact of adversarial training in the other tasks in ABSA namely Aspect Category Detection and Aspect Category Polarity will be investigated. 

\section*{Acknowledgment}
We would like to thank Adidas AG for funding this work.



\bibliographystyle{IEEEtran}
\bibliography{ref}
%\begin{thebibliography}{1}
%\end{thebibliography}

\end{document}